Pre-print Draft

# GENERATING DISPATCHING RULES FOR THE INTERRUPTING SWAP-ALLOWED BLOCKING JOB SHOP PROBLEM USING GRAPH NEURAL NETWORK AND REINFORCEMENT LEARNING

**Vivian Wen Hui Wong**[1]
Engineering Informatics Group,
Department of Civil and Environmental Engineering, Stanford University,
Stanford, CA, USA
e-mail: vwwong3@stanford.edu

**Sang Hun Kim**
Samsung Electronics Co., Ltd.,
Republic of Korea
e-mail: phd.kim@samsung.com

**Junyoung Park**
Department of Industrial and Systems Engineering,
Korea Advanced Institute of Science and Technology (KAIST),
Daejeon, Republic of Korea
e-mail: junyoungpark@kaist.ac.kr

**Jinkyoo Park**
Department of Industrial and Systems Engineering,
Korea Advanced Institute of Science and Technology (KAIST),
Daejeon, Republic of Korea
e-mail: jinkyoo.park@kaist.ac.kr

**Kincho H. Law**
Engineering Informatics Group,
Department of Civil and Environmental Engineering, Stanford University,
Stanford, CA, USA
e-mail: law@stanford.edu

## ABSTRACT

*The interrupting swap-allowed blocking job shop problem (ISBJSSP) is a complex scheduling problem that is able to model many manufacturing planning and logistics applications realistically by addressing both the lack of storage capacity and unforeseen production interruptions. Subjected to random disruptions due to machine malfunction or maintenance, industry production settings often choose to adopt dispatching rules to enable adaptive, real-time re-scheduling, rather than traditional methods that require costly re-computation on the new configuration every time the problem condition changes dynamically. To generate dispatching rules for the ISBJSSP problem, we introduce a dynamic disjunctive graph formulation characterized by nodes and edges subjected to continuous deletions and additions. This formulation enables the training of an adaptive scheduler utilizing graph neural networks and reinforcement learning. Furthermore, a simulator is developed to simulate interruption, swapping, and blocking in the ISBJSSP setting. Employing a set of reported benchmark instances, we conduct a detailed experimental study on ISBJSSP instances with a range of machine shutdown probabilities to show that the scheduling policies generated can outperform or are at least as competitive as existing dispatching rules with predetermined priority. This study shows that the ISBJSSP, which requires real-time adaptive solutions, can be scheduled efficiently with the proposed method when production interruptions occur with random machine shutdowns.*

## 1. INTRODUCTION

Effective scheduling strategies to various production scheduling problems have been widely studied in academia and industry with the goal to streamline manufacturing systems and hence to improve production efficiency. For example, the classical job shop scheduling problem (JSSP), which aims to find optimal assignment of jobs composed of operations given some prescribed machine sharing and precedence constraints, is an NP-hard combinatorial optimization problem that finds many practical applications. Many manufacturing systems in the real settings, however, have more constraints to consider than the capacity of machines. For example, many components of vehicles and machines are often expensive items that are huge in size. It is therefore not desirable to have to invest in the storage of intermediate components and products [1]. The lack of storage capacity is therefore a constraint in this case. Variations of the job shop problem without storage, often referred to as the blocking job shop problem, have diverse applications, including aircraft production [2] and steel manufacturing [3]. Developing effective scheduling strategies for these scenarios can help manufacturers optimize their production processes, leading to increased efficiency. The blocking job shop also finds applications in many areas beyond manufacturing, where logistics planning does not allow intermediate buffers. For example, the railway scheduling problem, where a train blocks a segment of the track until the train can be moved elsewhere, can be effectively modeled as a blocking job shop [4]. Furthermore, unforeseen interruptions to production, such as machine shutdowns, could occur that changes the list of available machines. To model modern manufacturing systems more realistically by considering both the *lack of storage capacities* and *production interruptions*, this work studies a new class of job scheduling problem, the interrupting swap-allowed blocking job shop scheduling problem (ISBJSSP).

[1] Corresponding author.

Many methods, such as mathematical optimization [5], branch-and-bound search [6] and meta-heuristic algorithms [7,8], have been developed to generate optimum or near-optimum solutions to the JSSP problems. However, these solutions are not adaptive, requiring a completely new execution when encountering a new scenario or a new configuration. These non-adaptive solutions are therefore not suitable for the ISBJSSP setting, where the problem condition constantly changes, for example, due to machine interruptions. To cope with potential dynamic changes, priority dispatching rules (PDRs), which are simple rule-based heuristics, are the most common approach used in modern manufacturing systems, as they can be applied instantaneously to an unknown instance. PDRs, first-in-first-out (FIFO) as an example, simply loads jobs based on some predetermined priority [9]. Although PDRs are widely used in real world situations due to their simplicity and speed, their performance varies widely depending on the problem condition. For example, shortest processing time (SPT) is a common benchmarking PDR that performs well in heavily loaded or congested job shop problem instances but fails with low load levels [10]. These simple rules, although they can deal with dynamic changes, have poor generalizability, and need to be manually selected or combined based on the job shop condition. Furthermore, with the random interruptions in the ISBJSSP formulation, where problem conditions change often, it is not clear apriori whether any of the PDRs can be effective on an ISBJSSP problem.

To improve the generalizability of dispatching rules, researchers have started to leverage artificial intelligence (AI) methods to solve job shop scheduling problems. To consider both the ability to adapt and to generalize, methods that are based on reinforcement learning (RL) are receiving increasing attention in the research community on planning problems. Much like PDRs, these methods output sequential decisions according to a dispatching policy. The difference is that rather than using predetermined priority rules, RL's dispatching policy is learned by observing and accumulating experience from previous simulations.

RL has been used to learn policies in various planning and scheduling problems. Traditional RL algorithms, such as Q-learning and its variants, are often used to learn dispatching rules for small-scale job scheduling problems with discrete state space [11,12]. In contrast, large-scale job shop scheduling problems are relatively unexplored. For large-scale, continuous state space problems, it is necessary to consider deep RL methods that approximate the value function.

Sparked by increased availability in computational power in recent years, deep RL methods combining deep neural networks with RL have received much attention due to its powerful ability to generate solutions for continuous state space. Examples of deep RL applications to planning and scheduling problems include task scheduling in computing [13], robotic scheduling for manufacturing [14], and semiconductor manufacturing scheduling [15]. However, the problem formulation of deep RL for job shop problems varies widely, for even the classical JSSP. Liu et al. [16] used process time matrices to represent the state space and trained a deep RL model to select from a list of existing PDRs. Park et al. [17] and Zhang et al. [18] use disjunctive graph and graph representation learning to obtain a vectorized state space, and directly learned new dispatching rules. For ISBJSSP, there lacks a formal Markov Decision Process formulation that enables the study of deep RL approach for this new class of job scheduling problem.

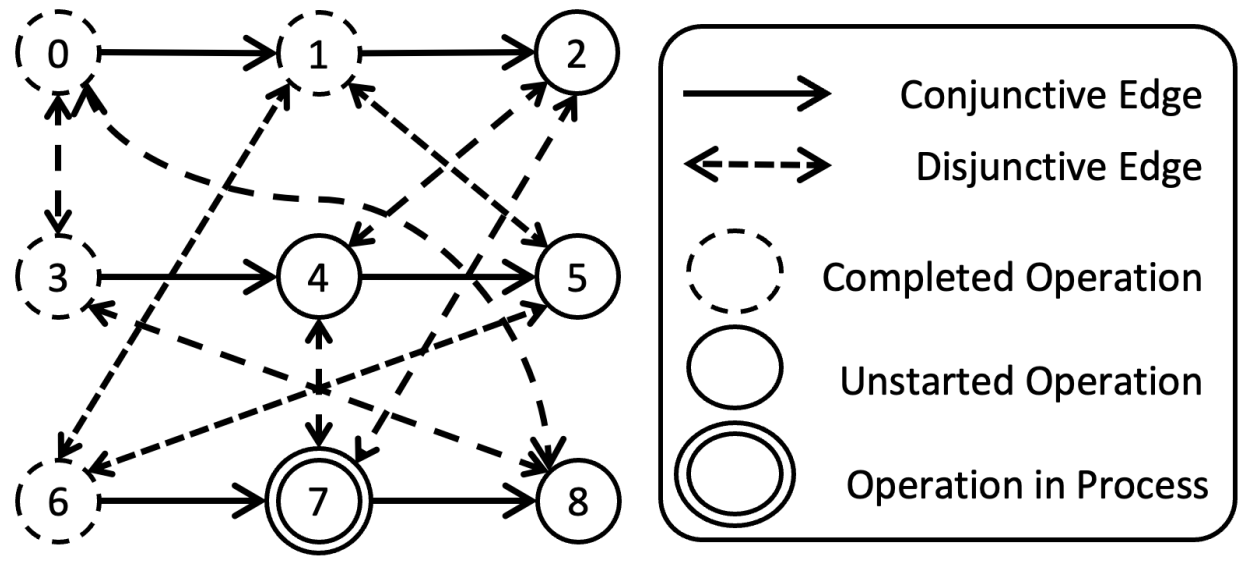

**Fig. 1 Example of a static disjunctive graph for an instance containing three jobs, each with three operations. Directed conjunctive edges represent precedence constraints. Bidirectional disjunctive edges represent machine constraints, where the nodes in a cycle are operations that require to be processed on the same machine. Nodes with dashed perimeters indicate completed operations. Nodes with solid perimeters indicate operations that have not been started. The double-outlined node indicates the operation currently being processed.**

This paper introduces the formulation of the ISBJSSP as a dynamic disjunctive graph, which serves as the state representation of the Markov Decision Process (MDP). To generate training data sets and experimental test scenarios, we implement an ISBJSSP simulator, building upon a Python-based JSSP simulator (pyjssp) previously developed by Park et al. [17]. The simulator is designed to simulate ISBJSSP instances to include blocking constraints, swapping conditions, and machine shutdown interruptions to mimic realistic concerns in practical applications. Using the simulator, GNN-RL scheduler models, combining graph neural network (GNN) and deep RL learning, are trained with randomly generated ISBJSSP instances. In this study, the performance of the trained models is evaluated on two sets of benchmark scenarios. The first test set is a set of $10 \times 10$ benchmark instances that are commonly used in job shop scheduling studies [19]. To demonstrate the scalability and generalization of the GNN-RL models, the second test set includes job shop instances with varying sizes [20]. The experimental results show that GNN-RL schedulers can be used to schedule unknown ISBJSSP instances robustly and efficiently and can potentially be applied in a real manufacturing environment without shutting down the entire job shop when interruption occurs.

The paper is organized as follows: Section 2 introduces the ISBJSSP formulation and how the problem can be modeled as a dynamic disjunctive graph and a Markov Decision Process. Section 3 describes the methodology employed to learn ISBJSSP scheduling models. Section 4 describes the experimental results obtained with benchmark ISBJSSP instances of different sizes. Section 5 concludes this paper with a brief summary and discussion.

## 2. PROBLEM FORMULATION

This section briefly introduces the background for the job shop problem and the various constraints that exist. The modeling of ISBJSSP as a dynamic disjunctive graph and a Markov Decision Process (MDP) is then described.

### 2.1 Job Shop Scheduling

For the classical JSSP of size $m \times n$, there exists a set of $n$ jobs, $O: \{O_1, O_2, \ldots, O_n\}$ , to be optimally allocated and executed on a set of $m$ machines, $M: \{M_1, M_2, \ldots, M_m\}$. Each job has a series of tasks or operations that must be processed according to the problem's precedence and machine-sharing constraints. In this study, without loss of generality, we assume each job, $O_j: \{o_{1j}, o_{2j}, \ldots, o_{pj}\}$ , has the same number of $p$ operations. Each operation $o_{ij}$ has a pre-defined processing time for completion. The precedence constraint implies that for all consecutive operations $o_{ij}$ and $o_{i+1,j}$ of job $O_j$ , $o_{ij}$ must be completed before starting $o_{i+1,j}$ . Furthermore, the machine-sharing constraint indicates that each operation $o_{ij}$ must be processed uninterrupted on a dedicated machine. Additionally, each operation of a job is assigned on a different machine, and each machine that is in process of an operation is only freed when that operation finishes. Therefore, given the above constraints, the number of machines $m = p$, the number of operations for each job. The objective of the classical JSSP is to find a schedule that minimizes the makespan, the total time to finish all jobs [21].

The classical job shop problem assumes that there is sufficient storage or buffer space available to store each job in between consecutive operations. However, buffers are undesirable in practical applications. Therefore, many real manufacturing applications are better modeled as the Blocking JSSP (BJSSP) [22]. BJSSP introduces the blocking constraint in that no buffers are available for storing a job as the job moves between machines; the job must wait on the machine until it can be processed on the next machine. That is, for any job $O_j$, its operation $o_{ij}$ is a blocking operation until $o_{ij}$'s succeeding operation, $o_{i+1,j}$, starts.

In practical job shops without buffers, the issue of deadlock is often mitigated through the implementation of blocking job swaps. The swapping mechanism is the repositioning of processed components across machines. Swapping cannot interrupt started but unfinished operations, and do not require the allocation of additional storage buffer. A deadlock situation occurs when no unprocessed operations could be processed, because each unprocessed operation is waiting for a blocked machine to become "unblocked" and available. Thus, BJSSP often allows swapping to avoid deadlocks, referred to as the swap-allowed blocking job shop problem (SBJSSP) [19]. A swap can be done if there exists a set of blocking operations, each one waiting for a machine blocked by another operation in the set. A swap resolves the deadlock and ensures that the manufacturing process can proceed and that there exists at least one solution to a randomly generated SBJSSP instance.

### 2.2 ISBJSSP

Although the SBJSSP models can be applied to many manufacturing production lines, there is an additional factor that exists in a real production line but is often overlooked - the possibility of production interruption, for example, caused by machine failures. While solution methods such as mathematical programming, branch and bound search and meta-heuristic methods (such as tabu search, genetic algorithm, simulated annealing, etc.) can generate optimal solution to static (uninterrupted) job shop scenarios, the dynamic scenarios with real time machine interruptions would require recomputing a new solution for each scenario change by the methods. The possibility of such interruption also results in priority dispatching rules [23] being generally favored in practice, as dispatching rules can easily adapt to dynamic changes in availability of machines in real time. Our work therefore includes this additional constraint where machine availability can be interrupted in the formulation: At any given time step, an idling machine in $M$ has a probability $P_{interrupt}$ that the machine is unavailable to process any job for a period of $T_{interrupt}$ time steps. If a job's next operation is waiting on an unavailable machine, the job will block the machine used by the precedent operation due to the lack of buffer. When the machine previously shut down becomes available again after $T_{interrupt}$ time steps, the machine will then process one of the waiting jobs, determined by the job shop scheduler. We refer the job shop problem with the interruption constraint as the interrupting swap-allowed blocking job shop problem (ISBJSSP).

### 2.3 Dynamic Disjunctive Graph Formulation

*2.3.1. Static Disjunctive Graph Representation of Job Shop*

In previous research focusing on uninterrupted problems like the JSSP and SBJSSP, it is a standard practice to employ a disjunctive graph, denoted as $G = (V, C \cup D)$ , as a representation for job shop problems [24]. Here, $V$ is the node set, with each node corresponding to an operation $o_{ij}$ of job $O_j$. $C$ is the set of conjunctive edges, where each edge connects two consecutive operations $o_{ij}$ and $o_{i+1,j}$ of job $O_j$. The conjunctive edges represent the set of processing order constraints. $D$ denotes the set of disjunctive edges, which connect two vertices if the two corresponding operations need to be processed on the same machine. The disjunctive edges represent the machine-sharing constraints. For uninterrupted job shop problems, the connectivity between nodes and edges remains static, resulting in a disjunctive graph that maintains the same topology throughout the entire duration of a given problem instance.

To illustrate how a static disjunctive graph represents an uninterrupted job shop problem, we provide a step-by-step example of a problem instance where swapping and blocking constraints exist, but interruptions do not yet occur. Figure 1 shows a static disjunctive graph of a small example instance. The instance contains three machines, on which three jobs, each with three operations, are to be processed. As an example, the first job contains the operations labeled with node numbers 0, 1, 2, which must be processed in this specified order due to the existence of precedence constraints. The precedence constraint is shown as directed edges in the disjunctive graph. Similarly, the second job must be processed in the order of 3, 4, 5, and the third job in the order of 6, 7, 8. The bi-directional disjunctive edges specify machine constraints. In our example, operations 1, 5, 6 need to be processed on a dedicated machine. Similarly, operations 0, 3, 8 share a machine, and operations 2, 4, 7 share a machine. At the time where this disjunctive graph was plotted as shown in Figure 1, operations 0, 1, 3, 6 are completed, and operation 7 is being processed. Furthermore, there is swap-allowed blocking to consider. For example, even when operation 7 is completed, it will block its machine, preventing operations 2 and 4 to be processed, until the part moves to the next machine to commence operation 8. At the current time step, all three machines are either blocked by an unstarted operation or busy processing an operation and are therefore not idle. As defined earlier, the machine shutdown interruption of probability $P_{interrupt}$ only occurs to idling machines.

To incorporate the time-dependent job shop information into the disjunctive graph with static topology, a node feature vector $x_v$ is assigned to each node $v = o_{ij} \in V$. We utilize the same node features as presented by Park et al. [17], which are stacked vectors with the following components:

- Node status: a one-hot index vector of size 3, indicating whether the operation $v$ is not yet started, being processed, or is completed.
- Processing time: the total time required to finish operation $v$.
- Degree of completion: the ratio of the accumulated processing time of $v$'s job to the total processing time of $v$'s job (i.e., the job that contains the operation).
- Number of succeeding operations: the number of operations including both $v$ and the operations after $v$ in $v$'s job.
- Waiting time: the time for which $v$ must wait for processing after it is ready to be processed.
- Remaining time: the remaining processing time needed to complete the operation $v$ once it has started.

*2.3.2. Dynamic Disjunctive Graph Representation of Job Shop*

In contrast to uninterrupted job shop problems, the ISBJSSP incorporates additional machine availability constraints in the form of random idling machine interruptions or failures. While we initialize all job shop instances (prior to any interruption) using the above-mentioned static disjunctive graph formulation, we model the randomly occuring interruptions using a dynamic disjunctive graph representation.

In the dynamic formulation, nodes and edges of the disjunctive graph are subjected to deletions and additions due to the machine availability constraint, making the graph connectivity constantly changing. More specifically, when a machine is shut down under probability $P_{interrupt}$, the nodes (i.e., the operations) that require the interrupted machine and their connected edges are temporarily removed from the graph, indicating that the machine is no longer observed at that time instance. When the machine recovers from its failure after $T_{interrupt}$ time steps, the previously removed nodes and edges are added back to the graph. Given the dynamic nature of the graph topology, we use $G_t$ to denote the disjunctive graph of an ISBJSSP instance at time $t$. The topology change is described here in more formal notations. Let $G_{t-1}$ be a dynamic disjunctive graph at time $t-1$ and $v$ a single node to be removed in the subsequent time step, time $t$. Concurrently, we denote an edge $c \sim v$ as a conjunctive edge incident to the node $v$, and $d \sim v$ as a disjunctive edge incident to the same node $v$. Upon the removal of node $v$, all conjunctive edges $c$ such that $c \sim v$ and disjunctive edges $d$ such that $d \sim v$ are simultaneously removed from graph, thus resulting in a distinct graph topology for $G_t$ compared to $G_{t-1}$. Likewise, upon the recovery of the machine associated with node $v$ at $G_{t+T_{interrupt}}$, we reinstate the node $v$ and the incident edges, reshaping the graph's topology once again.

We show the dynamic removal and addition of nodes and edges in an example in Figure 2. In this example, in the initial disjunctive graph snapshot of the ISBJSSP instance (Figure 2(a)), the dedicated machine for operation 1,5,6 is shut down before any operation could be processed. As a result, node 1,5,6 and all conjunctive and disjunctive edges incident to these nodes are removed from the graph (Figure 2(b)). After $T_{interrupt}$ time steps, we reinstate the removed nodes and edges, as shown in Figure 2(c). We note that this is a simplified illustration in which only one machine encounters shutdown and subsequent recovery. Consequently, the graph's topology is same as the initial graph upon recovery. In more complex scenarios, it is possible for multiple machines to undergo shutdown, either simultaneously or intermittently across the duration of the simulation, resulting in more complex changes in the dynamic graph's topology.

### 2.4 Markov Decision Process Formulation

The scheduling process of an ISBJSSP instance can be viewed as a sequential decision-making process. Specifically, ISBJSSP can be formulated as an MDP, denoted as a $(S, A, P, R, \gamma)$ tuple, whose elements represent the set of Markov states $(S)$, set of actions $(A)$, transition model $(P)$, reward function $(R)$ and discount factor $(\gamma)$, respectively.

- State: A disjunctive graph $G_t$ representing a snapshot of state $s_t \in S$ of the ISBJSSP instance at time $t$. $G_t$ here is the dynamic graph subjected to topology changes due to interruptions.
- Action: A scheduling action $a_t \in A$ of loading an operation to an available machine at time $t$. The action space of the ISBJSSP differs from that of the classical JSSP, wherein any machine not actively processing an operation is available. On contrary, in the case of SBJSSP and ISBJSSP, the inclusion of swapping and blocking necessitates an additional evaluation of machine availability status, explained later in Section 3.1.2.
- Transition model: The transition between states, which, in this study, is handled and generated by the ISBJSSP simulator developed in this study.
- Reward function: A function defined to stipulate the behavior of an action. The reward function used in this study mimics the utilization of a machine and is defined as

$$r_t = -n_{w_t} \quad (1)$$

where $n_{w_t}$ is the number of jobs waiting at time $t$.

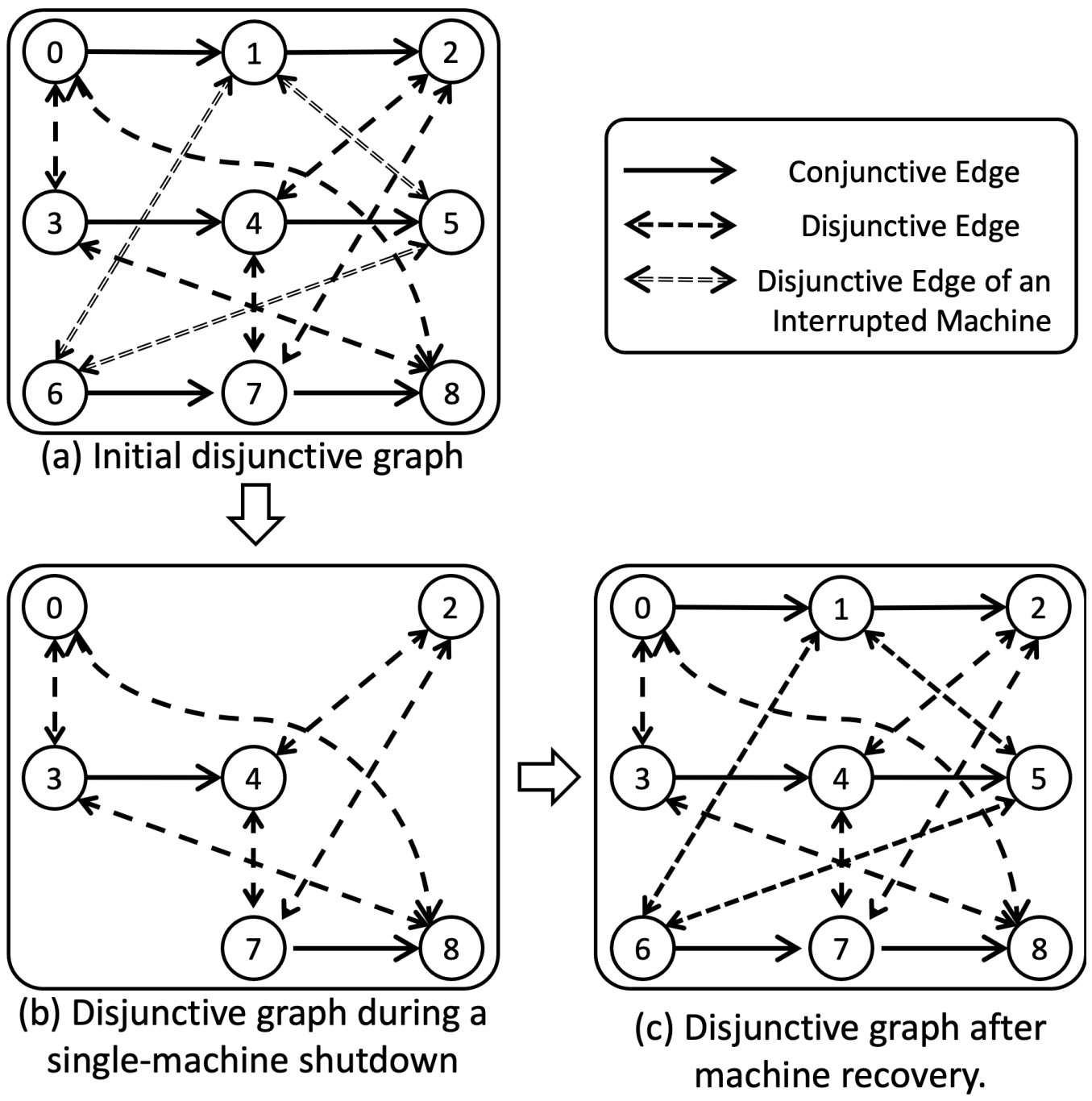

**Fig. 2 Example of a dynamic disjunctive graph. Nodes and edges are deleted when a machine is shut down, and are reinstated when the machine recovers.**

- Discount factor: The discount factor for "caring" of future reward by an action. In the ISBJSSP setting, the presence of random interruptions introduces greater uncertainty into future outcomes. Consequently, fine-tuning the discount factor is essential during model training to effectively adapt to these interruptions. We therefore compare two different values of the discount factor in our experimental results, presented later in Section 4.

## 3. METHODOLOGY

This section describes the workflow for deriving a policy to solve the ISBJSSP. The method consists of two main components: 1) the ISBJSSP simulator, which manages the dynamic graph generation and machine availability complexity, which are existent in the ISBJSSP setting but absent in the JSSP one, and 2) GNN-RL (graph neural network - reinforcement learning), a machine learning technique employed to learn the scheduling policy by observing the dynamic graphs. Figure 3 depicts the overarching framework of the proposed workflow.

A dynamic disjunctive graph $G_t$ (Figure 3(e)) is observed from the ISBJSSP simulator environment (Figure 3(d)) and is used as the input to a GNN model (Figure 3(f)) for representation learning. Differing from uninterrupted job shop problems, interruptions are additionally modeled by the previously discussed dynamic removal and addition of nodes and edges. The learned embedded graph $G_t'$ (Figure 3(a)), which captures such structural changes, is then used as the input to the RL algorithm, learning a parameterized policy, or a probability distribution of feasible actions, using an actor-critic model with proximal policy optimization (Figure 3(b,c)). Finally, an action to process a specific operation is sampled from the parameterized policy $\pi(\cdot\,|G_t)$ and executed via the ISBJSSP simulator (Figure 3(d)).

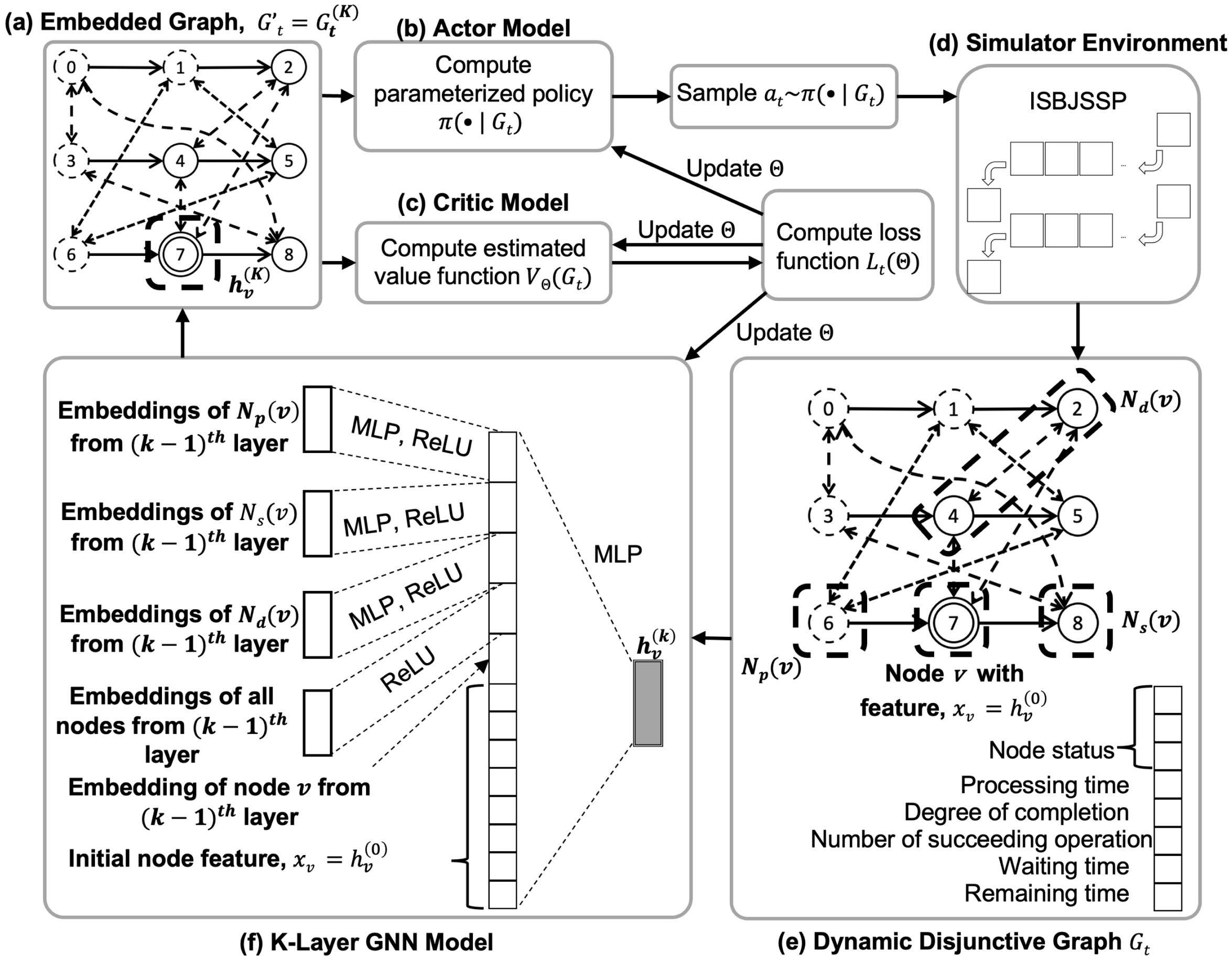

**Fig. 3 Proposed GNN-RL framework.**

In this section, we first discuss the building of the simulator, which must incorporate dynamic changes caused by interruptions, as well as swapping and blocking absent in classical JSSP simulators. Then, we present the GNN-RL procedure presented by Park et al. [17] for completeness of the methodology.

### 3.1 ISBJSSP Simulator

The primary objective of a simulator is to collect transition samples, $(G_{t-1}, a_{t-1}, r_{t-1}, G_t)$ . The generation of such transitions is problem specific, as different problem settings inevitably result in varying state and action spaces. Leveraging the JSSP simulator built by Park et al. [17], our modified simulator encompasses new features to facilitate the handling of machine interruptions, swapping, and blocking operations. These augmentations enable the simulator to generate transition samples following the ISBJSSP setting.

#### *3.1.1 Queue System to Manage Interruptions*

While interruptions of idling machines are modeled as the dynamic change in the disjunctive graph structure, there needs to be a system that manages and tracks the statuses of interrupted machines. In this subsection, we describe the use of queues in the simulator implementation to manage machine failures and recoveries.

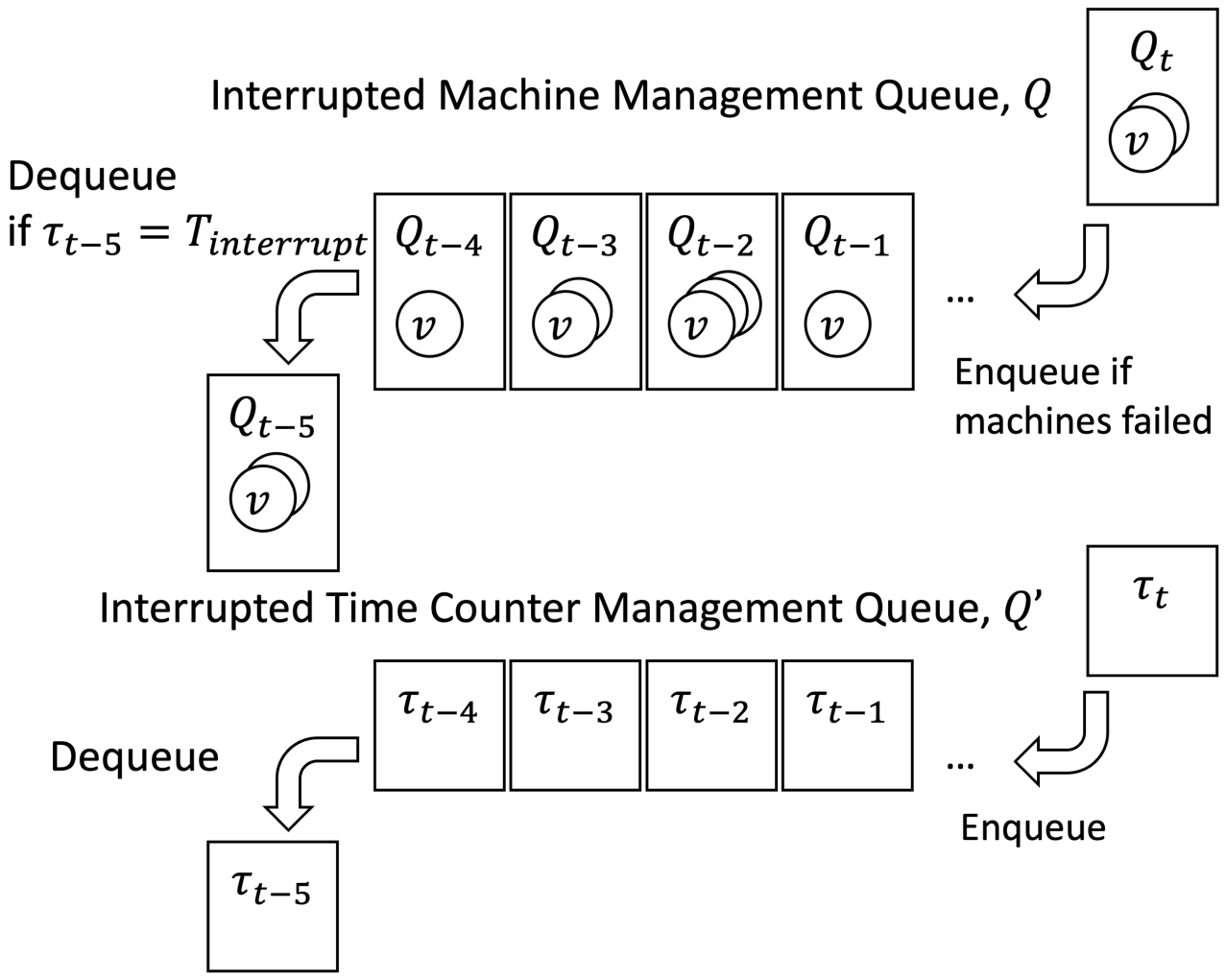

**Fig. 4 Queueing system for managing machine interruptions. Queue $Q$ stores removed nodes and Queue $Q'$ tracks shutdown times.**

Figure 4 illustrates an example of how a queue system is implemented to keep track of ongoing interruptions. To effectively manage the introduced machine interruptions, we uphold two queues. The first queue stores operations, each represented by a node, associated with the interrupted machines, denoted as $Q$. The second queue, labeled as $Q'$, tracks all interruption time counters. As the simulation proceeds and time step $t$ increases, operations associated to idling machines that encounter failures at time $t$ are stored in a node set $Q_t$, which is subsequently enqueued into queue $Q$. Additionally, an interruption time counter, $\tau_t$, that tallies the accumulated failure time since time step $t$ is also enqueued into the queue $Q'$. Upon the enqueueing of a node $v \in Q_t$ into queue $Q$, all conjunctive edges $c$ such that $c \sim v$ and disjunctive edges $d$ such that $d \sim v$ are removed from the dynamic graph $G_t$.

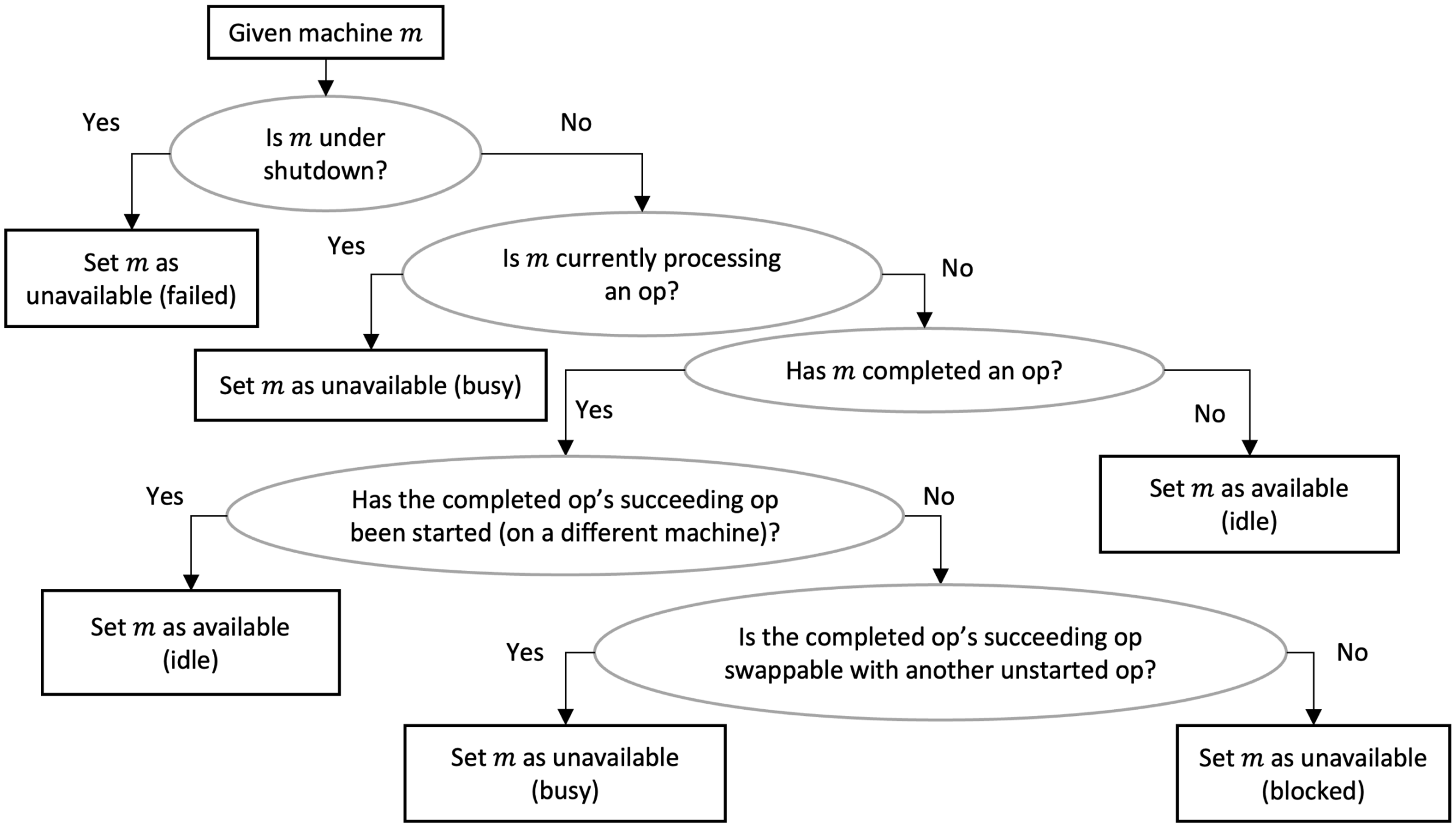

**Fig. 5 Schematic outline for assessing machine's availability status in ISBJSSP simulator.**

As machines recover from failure, indicated by the respective time counters reaching $T_{interrupt}$, the first elements within the respective queues are dequeued. Nodes from the dequeued node set and their corresponding conjunctive and disjunctive edges are then reinstated into the graph.

*3.1.2 Machine Availability Considering Interruption, Swapping and Blocking*

In our preceding discussion, we described the dynamic nature of the disjunctive graph states, namely the possible topology changes between subsequent graph snapshots, for example $G_{t-1}$ and $G_t$, due to machine interruptions. In addition to interruptions, the complexities of swapping and blocking operations also need to be addressed when building a simulator for the ISBJSSP. In JSSP simulators, $a_{t-1}$ is selected by loading operations from a list of available machines, which is simply identified as those not currently processing an operation. Yet, the ISBJSSP presents distinct considerations. Interrupted and blocked machines are rendered unavailable, and the introduction of swapping operations can unblock machines. Figure 5 outlines the algorithm to determine the availability status of a machine within the adapted ISBJSSP simulator framework.

Given a machine $m$, the objective of the algorithm presented in Figure 5 is to determine the status of $m$. The availability of $m$ is determined by:

- Case 1 – Unavailable (failed): The algorithm first checks whether the machine has been interrupted under the $P_{interrupt}$ shutdown probabilities. If the machine has been affected by the interruption, it is unavailable to process any operation.
- Case 2 – Unavailable (busy): If $m$ is currently processing an operation, say $o_{ij}$, then $m$ is unavailable to process any additional operation. Alternatively, if $m$ has finished processing $o_{ij}$ but $o_{i+1,j}$ is swappable with other unstarted operations, a swapped operation will immediately transfer to $m$, leaving $m$ unavailable to process any operation except the newly arrived operation.
- Case 3 – Available: If $m$ is not currently processing an operation and has never processed an operation at all, then $m$ is idle as it has not been put into work since initialization. Alternatively, if $m$ is not currently processing an operation and the succeeding operation ($o_{i+1,j}$) of the previously processed operation ($o_{ij}$) has been started, job $O_j$ has been transferred to another machine, rendering $m$ idle.
- Case 4 – Unavailable (blocked): If $m$ has processed $o_{ij}$, but $o_{i+1,j}$ is unable to start (due to its required machine being unavailable), the job $O_j$ cannot be transferred to another machine. $m$ is therefore deemed unavailable and blocked.

**3.2 Generating Scheduling Policy with GNN-RL**

In this section, the GNN-RL method (graph neural network – reinforcement learning) is presented below for completeness of the methodology. The discussion follows closely the notation of [17] and [27], and the GNN-RL procedure, which was designed for classical JSSP [17]. It should be emphasized, however, that the ISBJSSP requires the incorporation of the dynamic disjunctive graph, which is used as the state representation of the MDP and simulated by the modified ISBJSSP simulator.

*3.2.1 Representation Learning with GNN*

The goal of a GNN is to effectively capture and represent the structural information of the dynamic disjunctive graph, $G_t$, which serves as a snapshot of the dynamic job shop environment. While it is difficult to represent graph structures in the Euclidean space, GNN can be used to learn an embedded graph from $G_t$. The embedded graph contains an embedding vector for each node $v$ that represents the neighborhood connectivity and node feature information around the node. Note that the neighborhood connectivity here changes constantly as random interruptions occur, as shown previously in Figure 2. A GNN is a neural network that consists of layers of differentiable functions with learnable parameters and computes an embedding vector for each node in the graph. A GNN layer needs to be designed such that for each target node, the embedding of the target node is updated not only using the previous layer's target node embedding, but also node embedding aggregated from the neighboring nodes in order to learn embeddings that encode essential characteristics of $G_t = (V, C \cup D)$, such as the nodal features and the dynamic connectivity between jobs and machines. The learned embeddings serve as rich representations of the dynamic graph snapshot and is used as the input to the RL agent's decision-making process.

As shown in Figure 2(f), the computation process of a GNN layer implemented in this study can be separated into three steps. Firstly, a different multi-layer perceptron (MLP) network [25] with ReLU activation [26] is applied to each of the following three sets of nodes neighboring the target node $v$: the set of all precedent nodes $N_p(v)$ connected through the conjunctive (precedence constraint) edges, succeeding nodes $N_s(v)$ connected also through the conjunctive edges and disjunctive nodes $N_d(v)$ connected through the (bidirectional) disjunctive (machine-sharing constraint) edges. Unlike uninterrupted problems with static graph connectivities, $N_p(v)$, $N_s(v)$, $N_d(v)$ and the node set $V$ are time-variant, since interruptions could lead to deletion and addition of nodes within these neighborhoods. Secondly, the vector outputs of the three MLP networks, an aggregated representation of the overall graph, the node embedding updated in the previous layer, and the initial node feature of the target node are stacked in a vector, which, as the last step, is passed through another MLP network without activation. Mathematically, the operations of the $k^{th}$ layer of a GNN can be written as:

$$h_v^{(k)} = f_n^{(k)}\Big( \mathrm{ReLU}\big(f_p^{(k)}\big(\sum_{i\in N_p(v)} h_i^{(k-1)}\big)\big) \,\|\, \mathrm{ReLU}\big(f_s^{(k)}\big(\sum_{i\in N_s(v)} h_i^{(k-1)}\big)\big) \,\|\, \mathrm{ReLU}\big(f_d^{(k)}\big(\sum_{i\in N_d(v)} h_i^{(k-1)}\big)\big) \,\|\, \mathrm{ReLU}\big(\sum_{i\in V} h_i^{(k-1)}\big) \,\|\, h_v^{(k-1)} \,\|\, h_v^{(0)}\Big) \quad (2)$$

where $\mathrm{ReLU}(\cdot) = \max(0,\cdot)$ is a non-linear activation function. The previously mentioned MLP networks are denoted by $f_p, f_s,$ and $f_d$, each computes a vector from node embeddings in the neighborhood of $v$ (i.e., $N_p(v), N_s(v),$ and $N_d(v)$, respectively). $V$ is the set of all nodes of the graph $G_t = (V, C \cup D)$. $h_v^{(0)}$ is the feature vector $x_v$ of node $v$. $\|$ is the vector concatenation operator.

After $K$ GNN layers, we have computed an embedded graph $G_t^{(K)}$, as shown in Figure 2(a), whose node features are now the updated embedding vectors $h_v^{(K)}\,\forall\, v \in V$, from the input disjunctive graph $G_t$ in Figure 2(e) with initial node features $h_v^{(0)}$.

*3.2.2 Dispatching Policy Learning with RL*

Outputted by GNN described in the previous section, the graph embedding $G_t^{(K)}$, a rich representation of both dynamic changes in neighborhood connectivity due to the presence of interruptions and time-dependent nodal features, is used as the input for the RL algorithm. More specifically, an actor-critic method is used [27]. As the name suggests, there are two neural networks in the RL process: an actor and a critic. As shown in Figure 2(b), the actor $\pi(a_t^v|G_t^{(K)})$ maps the embedded graph $G_t^{(K)}$ to the probability distribution over the set of all available actions, or the set of processible nodes. The actor model, used to compute the parameterized policy, is structured like a softmax function computing the probability of performing action $a_t^v$ for the current state $G_t^{(K)}$ as follows:

$$\pi(a_t^v|G_t^{(K)}) = \frac{\exp\left(f_l\left(h_v^{(K)}\right)\right)}{\sum_{u\in A_{G_t}} \exp\left(f_l\left(h_u^{(K)}\right)\right)} \quad (3)$$

where $a_t^v$ denotes the action of selecting operation (node) $v$ to process and $v$ is a processible node in the disjunctive graph's node set $V$. Following the same notation in the previous section, $h_v^{(K)}$ is the embedded vector of node $v$, and $f_l$ $(l = p, s, d)$ denotes a MLP network. $A_{G_t}$ represents the set of available actions for the disjunctive graph $G_t$. $A_{G_t}$ is obtained by querying a list of available machines, as discussed by the algorithm previously shown in Figure 5, due to the additional problem constraints imposed by the ISBJSSP.

The critic model, as depicted in Figure 2(c), is another network that learns the value function to reliably optimize the policy. The current study approximates the critic function as

$$V^{\pi}\left(G_t^{(K)}\right) \approx f_v\left(\sum_{i\in V} h_i^{(K)}\right) \quad (4)$$

where $f_v$ is an MLP network, and $\sum_{i\in V} h_i^{(K)}$ returns a summation of all node embeddings.

It can be observed that the policy $\pi(a_t^v|G_t^{(K)})$ learned by the actor is now parameterized by a set of parameters $\Theta = \{\theta_p, \theta_s, \theta_d, \theta_n, \theta_l, \theta_v\}$, corresponding to the MLP networks $f_p, f_s, f_d, f_n, f_l$ and $f_v$. The parameters can be iteratively updated via gradient ascent. In each training iteration, we use the policy $\pi_{\Theta_{old}}$ with the current "old" parameters $\Theta_{old}$ to interact with the job shop simulator (Figure 2(d)) and collect transition samples. The parameters $\Theta$ are updated to optimize the policy:

$$\Theta = \Theta_{old} + \eta \nabla_{\Theta} L(\Theta) \quad (5)$$

where $\eta$ is the learning rate and $L(\Theta)$ denotes an objective function to be optimized for an optimal policy.

In this work, proximal policy optimization (PPO) is employed to optimize the policy. To prevent unstable training due to substantial policy changes and encourage exploration during training, Schulman et.al. [27] proposes an objective function $L(\Theta)$ to be optimized at each time step $t$ as follows:

$$L_t(\Theta) = \mathbb{E}[L_t^{CLIP}(\Theta) - \alpha L_t^{VF}(\Theta) + \beta E_t(\pi_\Theta)] \quad (6)$$

where $\alpha$ and $\beta$ are parameters for the objective function. $L_t^{CLIP}(\Theta), L_t^{VF}(\Theta)$, and $E_t(\pi_\Theta)$ are, respectively, a clipped-surrogate function, square-error value function loss, and an entropy bonus, which are given as follows [23]:

1. The clipped-surrogate function is defined as:

   $$L_t^{CLIP}(\Theta) = \mathbb{E}\left[\min(\rho_t \mathcal{A}_t, clip(\rho_t, 1-\epsilon, 1+\epsilon)\mathcal{A}_t)\right] \quad (7)$$

   where $\rho_t$ denotes a probability ratio of the current and old policies as

   $$\rho_t = \frac{\pi_\Theta\left(a_t|G_t^{(K)}\right)}{\pi_{\Theta_{old}}\left(a_t|G_t^{(K)}\right)} \quad (8)$$

   and the estimator of the advantage function $\mathcal{A}_t$ at time step $t$ is computed as:

   $$\mathcal{A}_t = \delta_t + (\gamma\lambda)\delta_{t+1} + \cdots + (\gamma\lambda)^{T-t+1}\delta_{T-1} \quad (9)$$

   $$\text{and } \delta_t = \rho_t + \gamma V_\Theta\left(G_{t+1}^{(K)}\right) - V_\Theta\left(G_t^{(K)}\right) \quad (10)$$

   The coefficients $\gamma$ and $\lambda$ are, respectively, the discount factor and the parameter for the advantage function estimator. The clip operation ensures that $\rho_t$ does not move outside the interval $[1-\epsilon, 1+\epsilon]$ , thereby preventing substantial changes in policy.

2. The square-error value function loss is given as:

$$L_t^{VF}(\Theta) = \left(V_\Theta\left(G_t^{(K)}\right) - V_t^{target}\right)^2 \quad (11)$$

\

where $V_t^{target} = \sum_{i=t}^{T} r_i$ denotes the sum of rewards.

3. The entropy bonus term for the current policy $\pi_\Theta(a)$ is introduced to ensure sufficient exploration and is defined as:

$$E_t(\pi_\Theta) = -\sum_a \log(\pi_\Theta(a))\, \pi_\Theta(a) \quad (12)$$

where $a$ is an action performed given the current embedded graph $G_t^{(K)}$.

The PPO procedure maximizes the objective function $L(\Theta)$ by updating the parameters $\Theta$ following the gradient direction $\nabla_\Theta L(\Theta)$. Further discussion of the PPO algorithms can be found in References [27] and [17]. It should be emphasized here that $G_t^{(K)}$ is learned from $G_t$, which is subjected to dynamic node and edge removal and addition due to the interruptive nature inherent to the ISBJSSP. The graph topology therefore could stochastically change during training, leading to variations between iterations even when trained on the same problem instance.

## 4. EXPERIMENTAL RESULTS

This section describes the results of the experiments on a number of benchmark instances to evaluate the schedulers trained using GNN-RL. We will first describe the benchmark instances and details of the schedulers used in the experiments. We then report the experimental results, including: (1) a performance comparison of the GNN-RL method with other dispatching rules for the standard SBJSSP (a special case of the ISBJSSP with $P_{interrupt} = 0$); (2) a performance comparison demonstrating the practicability of the above methods for the ISBJSSP, which is subjected to random machine interruptions with probability $P_{interrupt} > 0$; and (3) a demonstration of the GNN-RL method's ability to generalize a model trained with instances of a specific size to handle ISBJSSP instances of different sizes.

### 4.1 The Baseline Benchmark Problem Instances

As a baseline to evaluate and compare the GNN-RL methodology with the PDR methods, a set of 18 job shop scheduling problem instances, each being $10 \times 10$ in size (consisting of 10 machines and 10 jobs), are employed. Each job involves 10 operations and, depending on the benchmark instance, the operations may have different machine processing time. The 18 instances are commonly used as benchmarks for job shop scheduling [19]. Even though this study focuses on the ISBJSSP where machine interruptions may occur, the benchmark instances serve as a fair metric for evaluating scheduling efficacy between the GNN-RL method and the PDR methods.

### 4.2 Scheduler Models and Configurations

*4.2.1 Priority Dispatching Rules (PDRs)*

As mentioned before, PDRs [23] are the most common approaches employed in practice for generating immediate solutions for scheduling job shops with unseen instances. We therefore compare the makespans obtained by the GNN-RL schedulers with those obtained using the following PDRs for prioritizing the preference for job execution:

- Most Total Work Remaining (MTWR): the job that has the greatest number of remaining operations
- Least Total Work Remaining (LTWR): the job that has the fewest number of remaining operations
- Shortest Processing Time (SPT): the job whose next operation has the shortest processing time
- Longest Processing Time (LPT): the job whose next operation has the longest processing time
- First In First Out (FIFO): the first job that arrives
- Last In First Out (LIFO): the last job that arrives
- Shortest Queue Next Operation (SQNO): the job whose next operation requires a machine that has the fewest number of jobs waiting
- Longest Queue Next Operation (LQNO): the job whose next operation requires a machine that has the most number of jobs waiting
- Shortest Total Processing Time (STPT): the job with the shortest total processing time
- Longest Total Processing Time (LTPT): the job with the longest total processing time
- Random: the job that is randomly selected from the set of all doable jobs

The PDRs can be applied irrespective of whether machine interruptions occur during the job operations.

*4.2.2 GNN-RL Schedulers*

Initially targeted for the baseline benchmark instances of $10 \times 10$ in size, a random ISBJSSP instance of size $m \sim \mathcal{U}(5,9) \times n \sim \mathcal{U}(m,9)$ and operation processing times from $\mathcal{U}(1,99)$, where $\mathcal{U}$ denotes a uniform distribution, is generated using the ISBJSSP simulator for training the GNN-RL models. The order of machines that each job visits is randomly permuted. After 20 episodes of training on the instance, a new ISBJSSP instance, once again with size $m \sim \mathcal{U}(5,9) \times n \sim \mathcal{U}(m,9)$, processing times from $\mathcal{U}(1,99)$ and randomly permuted machine order, is generated every 100 iterations. Note that, as discussed in a latter section, even though training is conducted on small-size instances to limit computational time and demand, the scheduling strategy that the model learns can be transferred to solve instances of other sizes effectively. Algorithm 1 outlines the procedure to train a GNN-RL scheduler. In each training iteration, we collect transition samples $(G_{t-1}, a_{t-1}, r_{t-1}, G_t)$ spanning $T$ timesteps, equivalent to a full simulation of the initialized ISBJSSP instance. Throughout the simulation, machine failures, with a probability of $P_{interrupt}$, and downtimes of $T_{interrupt}$ occur. These interruptions are effectively managed using the modified ISBJSSP simulator using the queue system and the machine availability determining algorithm. Subsequently, the collected transition samples are utilized to optimize the objective function $L_t(\Theta)$ through the PPO method.

An Adam optimizer [28] with a learning rate ( $\eta$ ) of $2.5 \times 10^{-4}$ is used. Two different discount factors ($\gamma$) are used in training, which are 0.9 and 1.0 (no discount) respectively. We use a GNN with $K = 3$ layers to obtain the graph embeddings. The MLP networks, namely $f_p$, $f_s, f_d, f_n, f_l$ and $f_v$, each consists of two hidden layers with 256 ReLU activation units. $f_p, f_s$ and $f_d$ have 8-dimensional inputs and outputs. $f_n$ has 48-dimensional input and 8-dimensional output. $f_l$ and $f_v$ have 8-dimensional inputs and scalar outputs. For the PPO hyperparameters, we set $\lambda = 0.95$, $\epsilon = 0.2$, $\alpha = 0.5$, and $\beta = 0.01$, which are the same as proposed in [17].

For comparison purpose, models are trained without interruptions and with the possibility of machine interruptions. When trained on SBJSSP (without interruptions), we set the probability of interruption $P_{interrupt} = 0$. The trained GNN_RL models are used for a baseline comparison with the PDR methods. For models that are trained on ISBJSSP instances, we train a different model for each $P_{interrupt}$ value, ranging from 1% to 20%. The trained GNN-RL models for the ISBJSSP are then used to compare with the SPT (shortest process time) priority rule (which achieves the shortest makespan among the PDRs for the non-interrupting SBJSSP) for the cases with machine interruptions. All experiments are conducted on a machine equipped with an Intel Core i7-7820X processor.

### 4.3 Results on Non-interrupting SBJSSP

The goal of the job shop problem is to minimize the makespan, which is employed here as the evaluation criterion for comparing the performances of the different SBJSSP schedulers. We perform hyperparameter tuning on the discount factor, an important hyperparameter dictating the importance of future rewards, which could be affected by the new problem constraints posed by the ISBJSSP. We report two GNN-RL models with different discount factor values, namely GNN-RL (1) with $\gamma = 0.9$ and GNN-RL (2) with $\gamma = 1.0$. Figure 6 shows the results of the two GNN-RL schedulers and the makespans obtained using the PDR schedulers for the 18 benchmark instances. Among the PDRs, the best scheduler appears to be problem dependent. As shown in Figure 6, on most of the benchmark instances, at least one of the GNN-RL schedulers is able to outperform or as competitive as the PDR schedulers, assuming no machine interruptions occur.

Table 1 reports the sum of the makespans of all problem instances. As shown, on average over all the benchmark instances, the GNN-RL schedulers produce shorter makespans than those by the PDRs schedulers. The GNN-RL (2) scheduler with $\gamma = 1.0$ has the best average performance. Among the PDR schedulers, the SPT strategy, which prioritizes the jobs according to the next operation having the shortest processing time, appears to perform the best on the average.

Table 1 also summarizes the total validation times over all 18 benchmark instances for each scheduler. Note that unlike most batch-based deep learning tasks, the speed on the validation results depends heavily on the central processing unit (CPU)'s

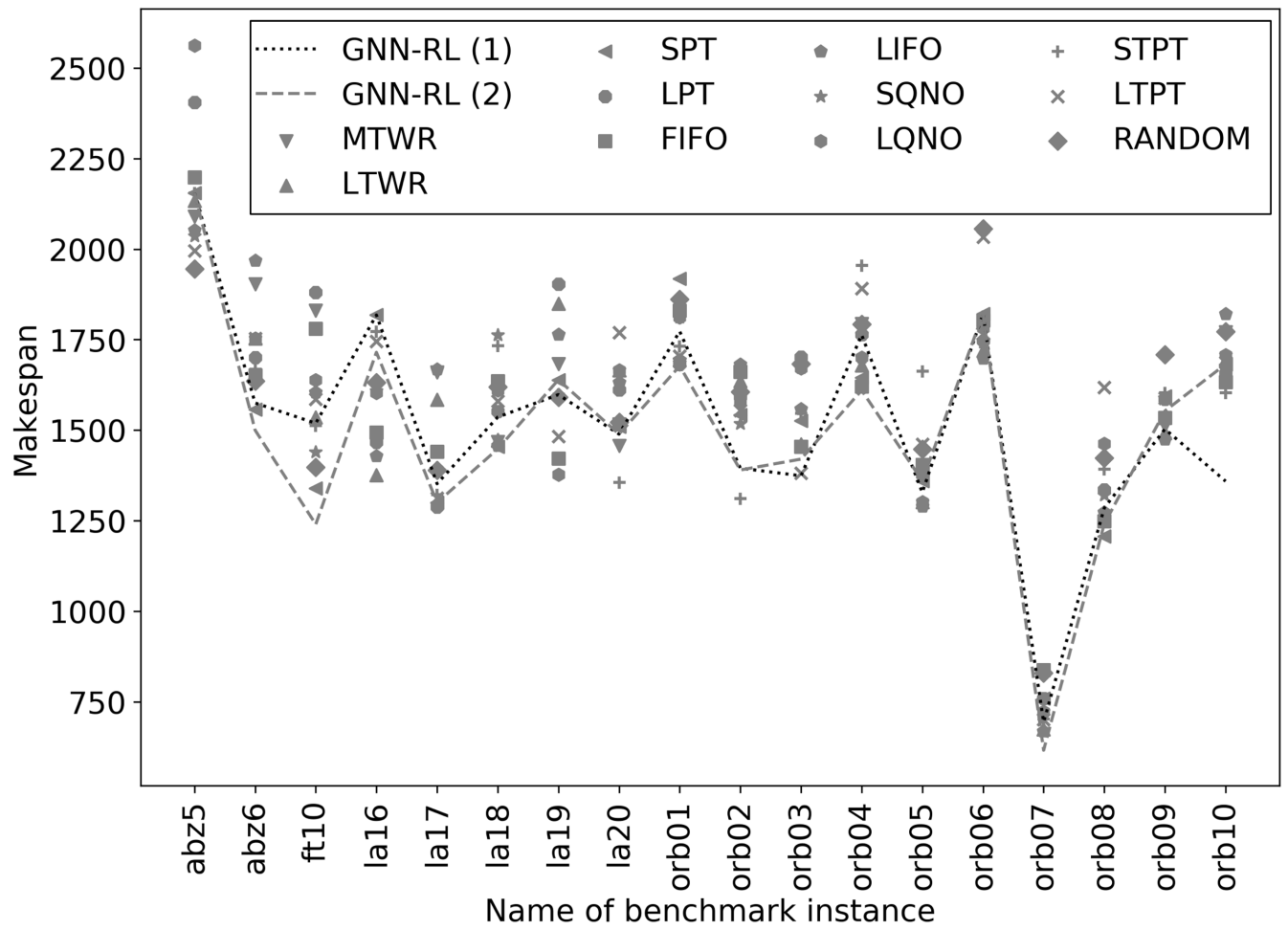

**Fig. 6 Makespans obtained using the two trained GNN-RL schedulers and the PDRs on the 18 benchmark instances without machine interruptions.**

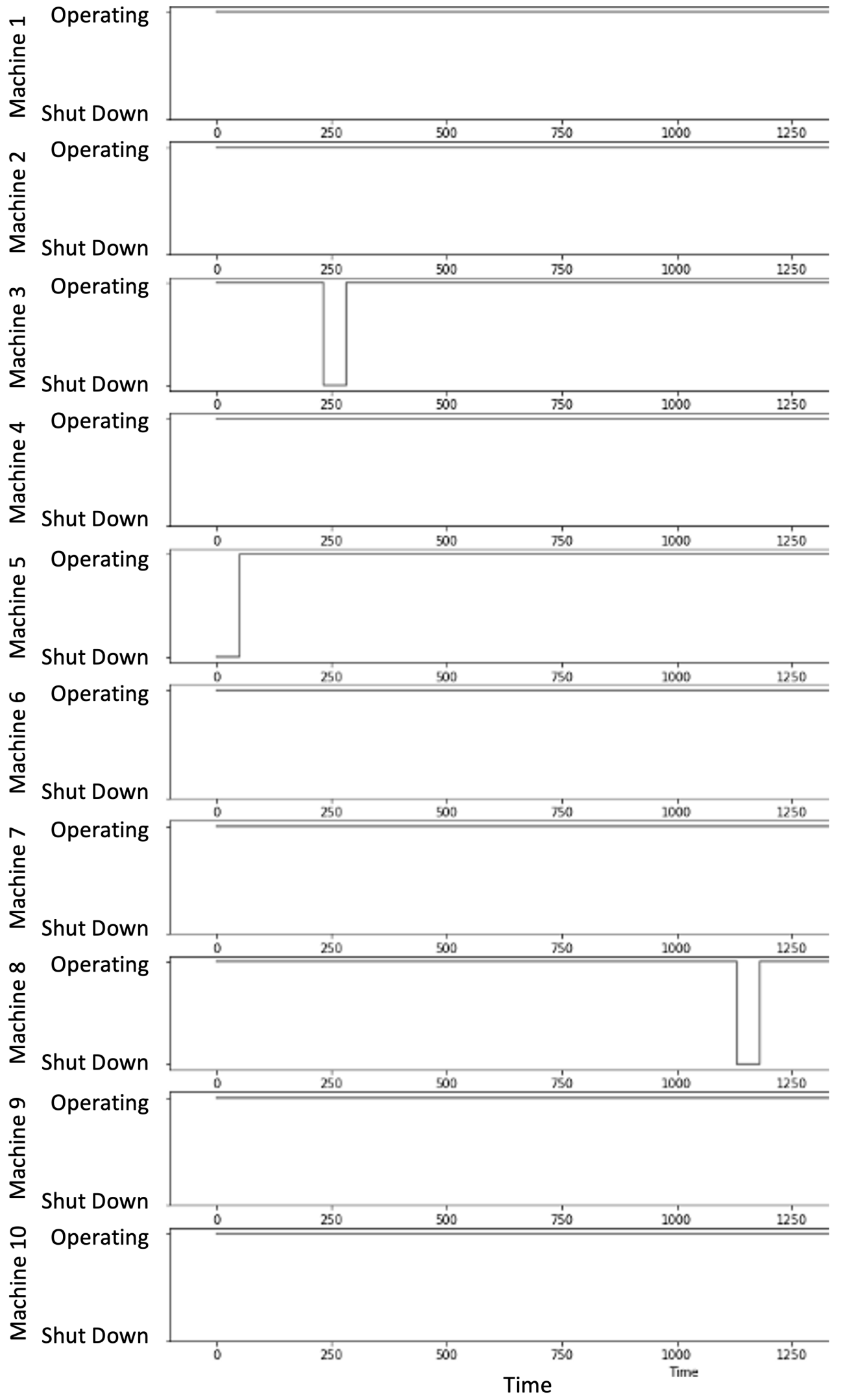

**Fig. 7 Status of 10 machines for an interrupted scenario. In this example, each machine, when idling, has a $P_{interrupt} = 5\%$ probability of shutting down for $T_{interrupt}$ =50 time steps.**

**Algorithm 1** Training procedure for GNN-RL scheduler

Generate a random ISBJSSP instance as starting state $G_0 = (V, C \cup D)$;
Initialize parameters $\Theta$ and the parameterized policy $\pi_\Theta$;
Initialize iteration = 0;
**repeat**
  iteration $+= 1$;
  **for** step $t = 1,2, \dots, T$ **do**
    **for** each node $v = o_{ij} \in V$ **do**
      **if** $o_{ij}$'s machine is idle **then**
        Add $v$ to $B_t$ with probability $P_{interrupt}$ ;
        Remove each node $v$, edges $c \sim v$ and $d \sim v$ from the graph $G_t$;
        Initialize interruption time counter $q_t$= 0;
        Enqueue $B_t$ into a queue of node sets $B$;
        Enqueue $q_t$ into a queue of counters $Q$;
      **end if**
    **end for**
    Denote $t = t - T_{interrupt}$;
    **if** the oldest element of $Q$ is $q_t = T_{interrupt}$ **then**
      Dequeue $q_t$ from queue $Q$;
      Reinstate all nodes $v$ such that $v \in B_t$ back into the graph $G_t$;
      Reinstate all edges $c$ such that $c \sim v$ and $d$ such that $d \sim v$ back into the graph $G_t$;
    **end if**
    Add 1 to all elements $q \in Q$;
    Observe and collect transition sample $(G_{t-1}, a_{t-1}, r_{t-1}, G_t)$;
    Query list of available machines (see Figure 5);
    Execute action $a_t \sim \pi_\Theta\left(\cdot \mid G_t^{(K)}\right)$ to assign operations to queried list of available machines;
  **end for**
  Update parameters $\Theta$ with gradient ascent to maximize $L_t(\Theta)$, calculated from the collected transition samples;
  **if** iteration = 100 **then**
    Generate a new random ISBJSSP as starting state $G_0$;
    Reset iteration = 0;
  **end if**
**until** validation performance has converged.

speed, which drives the simulation speed of our ISBJSSP simulator. As expected, the GNN-RL schedulers take slightly longer to compute solutions compared to PDR schedulers due to the fact that the scheduling action is computed from a policy, rather than derived a simple handcrafted rule like the PDR schedulers. The choice of scheduler will ultimately depend on the specific requirements of the task and the available computational resources.

### 4.4 Real-time Adaptive Scheduling of the ISBJSSP for the Baseline Benchmark

In practice, unforeseen interruptions could occur during production. For example, machines in a production line can misbehave unexpectedly at times that require a shutdown. To assess whether the GNN-RL method can cope with real-time changes, we simulate the scenarios where at any given time step, each idling machine, excluding those in the middle of processing a job, has a certain probability of failing or shutdown, denoted as $P_{interrupt}$, for a duration of $T_{interrupt} = 50$ time steps. During a simulated machine failure, no further job can be assigned to the machine. More specifically, a machine being shutdown is equivalent to removing the associated nodes and edges from the disjunctive graph representation for 50 time steps according to our problem formulation. The schedulers have no prior knowledge on the probability and the down time of the machines. As an example of the machine shutdown schedule given interruptions, Figure 7 shows a single simulation of 10 machines operating with a probability of interruption $P_{interrupt}$ of 5% and a $T_{interrupt}$ of 50 time steps.

**Table 1 Total makespan computed by GNN-RL and PDR schedulers for 18 benchmark instances, without machine interruptions.**

| Scheduler name | Discount factor | Total makespan | Time to validate 18 benchmark instances (s) |
|---|---|---|---|
| GNN-RL (1) | 0.9 | 27337 | 7.76 |
| GNN-RL (2) | 1.0 | **26856** | 7.52 |
| MTWR | - | 28687 | 5.24 |
| LTWR | - | 27996 | 5.53 |
| SPT | - | 27827 | 5.40 |
| LPT | - | 29009 | 5.52 |
| FIFO | - | 28169 | 5.29 |
| LIFO | - | 28497 | 5.57 |
| SQNO | - | 28563 | 5.39 |
| LQNO | - | 28818 | 5.43 |
| STPT | - | 28439 | 5.42 |
| LTPT | - | 28845 | 5.42 |
| RANDOM | - | 28988 | 5.38 |

Three GNN-RL models are trained. They include the same two models, namely GNN-RL (1) and GNN-RL (2) trained without machine interruptions, as described in Section 4.3. The third model, GNN-RL (3), has the same hyperparameters as GNN-RL (2), but is trained with the same probability of interruption, $P_{interrupt}$, as assigned to the simulation scenario.

We perform 50 simulations for each of the benchmark instances for a number of $P_{interrupt}$ values, ranging from 1% to 20%. Among the PDR schedulers, SPT shows the best average performance for almost all the cases and is therefore employed here to compare with the GNN-RL schedulers. Figure 8 shows a comparison of the average results between the GNN-RL and the SPT schedulers. It can be seen that for most instances, the GNN-RL schedulers outperform or are as competitive as the SPT scheduler for $P_{interrupt} < 10\%$. Furthermore, the GNN-RL (2) model and the GNN-RL (3) model trained with interruptions perform consistently better than the GNN-RL (1) model. Moreover, Figure 5 plots the performance, averaged over the 18 benchmark instances, of each scheduler with respect to $P_{interrupt}$. Also shown in Figure 9 are the means and standard deviations (Std) of the scheduling results for 50 randomly generated instances. Based on the results shown in Figures 8 and 9, the following can be observed:

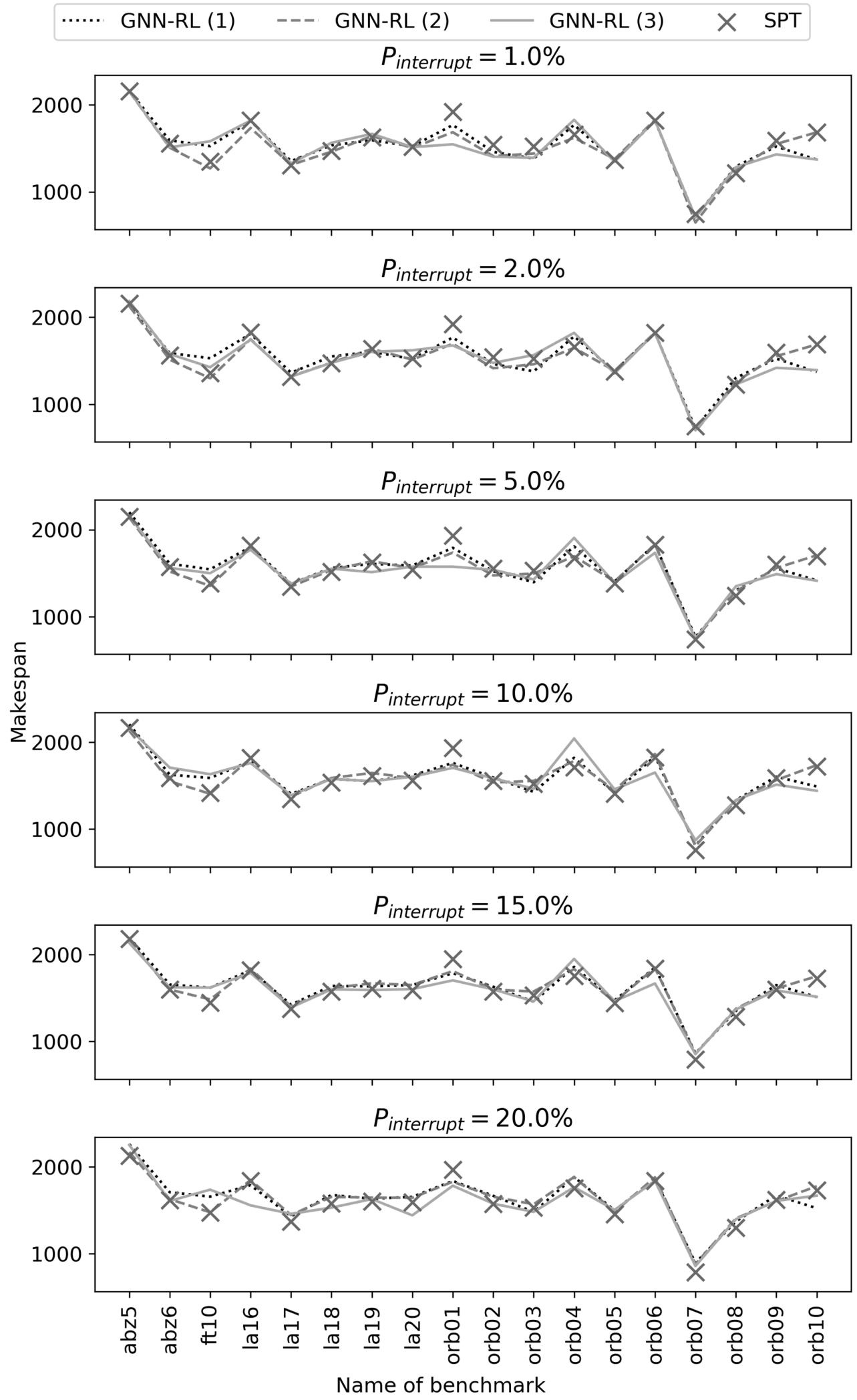

**Fig. 8 Mean of makespans using GNN-RL and SPT schedulers on the 18 baseline benchmark instances.**

1. As expected, when the probability of interruption for the machines increases, the makespans produced by the schedulers for completing all the jobs increase. This is likely due to the fact that machine interruptions reduce machine availability, as shown in Figure 5.
2. All GNN-RL models produce more efficient makespans than the SPT scheduler when the probability of machine interruption is lower than 5%. It can be seen from Figure 9 that, with interpolation, the GNN-RL models trained without machine interruptions can potentially be effective up to 8-10% probability of machine interruptions. Beyond $P_{interrupt} = 10\%$, the SPT scheduler produces more efficient makespans in this case study.
3. It is interesting to observe that, for the set of benchmark instances tested, the GNN-RL (3) model trained with the same probability of interruption assigned to the simulator performs quite competitively for almost all cases.
4. As can be seen in Figure 9, when the probability of interruption becomes high ($P_{interrupt} > 10\%$), the standard deviations for the GNN-RL schedulers are higher than the SPT scheduler. The higher standard deviation is probably due to the increase in uncertainties on machine interruptions that affect the predictability of the trained GNN-RL models.

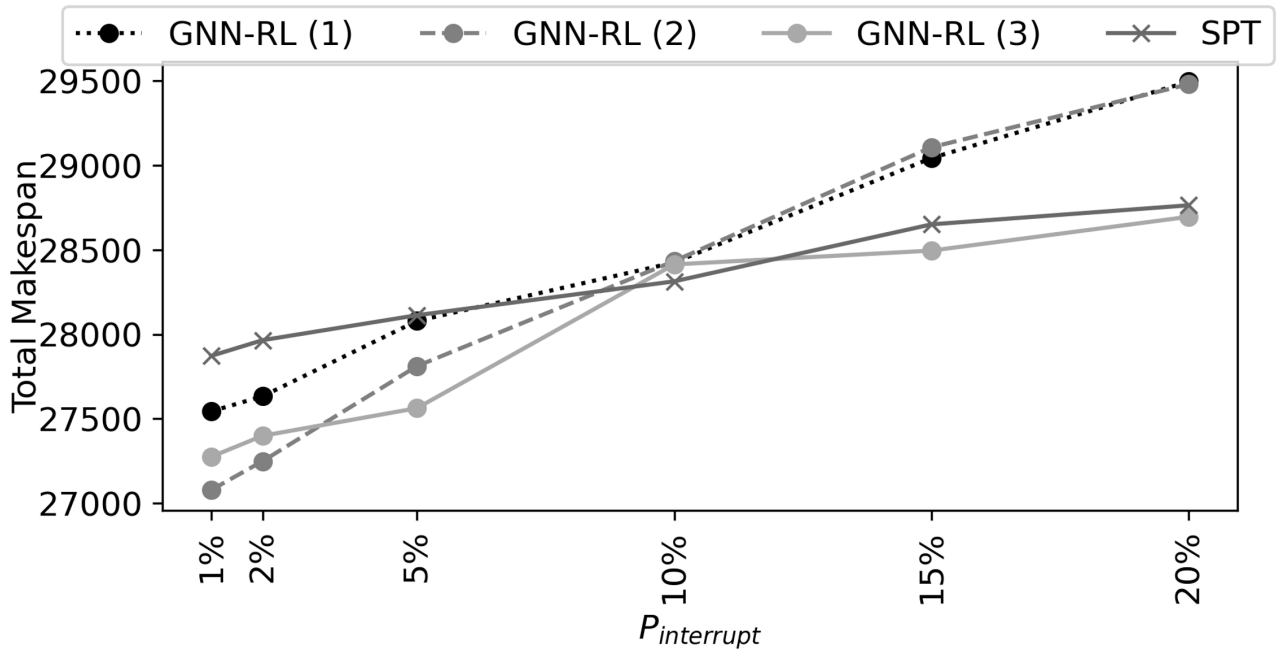

| | | Total Makespan | | | | | |
|---|---|---|---|---|---|---|---|
| $P_{interrupt}$ | | 1% | 2% | 5% | 10% | 15% | 20% |
| GNN-RL (1) | Mean | 27543 | 27633 | 28079 | 28425 | 29044 | 29496 |
| | Std | 199.2 | 312.9 | 415.8 | 451.8 | 487.1 | 561.2 |
| GNN-RL (2) | Mean | **27077** | **27248** | 27811 | 28433 | 29108 | 29481 |
| | Std | 188.5 | 247.7 | 357.8 | 486.2 | 487.6 | 500.6 |
| GNN-RL (3) | Mean | 27274 | 27398 | **27562** | 28414 | **28496** | **28696** |
| | Std | 174.5 | 242.4 | 363.4 | 493.5 | 474.4 | 509.7 |
| SPT | Mean | 27872 | 27965 | 28113 | **28313** | 28651 | 28764 |
| | Std | **142.4** | **177.5** | **239.6** | **339.9** | **402.2** | **332.4** |

**Fig. 9 Total makespans of GNN-RL and SPT schedulers for different probabilities of interruption.**

In summary, based on the experimentation on the 18 benchmark instances, the GNN-RL schedulers are shown to be robust for the scenarios where the probability of interruptions for each machine is less than 10%, even when the GNN-RL model is trained based on the scenarios with no machine interruptions.

### 4.5 Scheduling ISBJSSP Instances of Difference Sizes

To assess the scalability and generalization of GNN-RL models to instances of different sizes, we apply the same GNN-RL models trained previously with job shop instances of size $m \sim \mathcal{U}(5,9) \times n \sim \mathcal{U}(m, 9)$ to the 40 LA benchmark instances, with a range of sizes from 10×5 to 30×10 and 15×15 [20]. Makespans are obtained for 50 simulations on each of benchmark instances. Figure 10 shows the makespans computed with the SPT scheduler, the GNN-RL (1) and (2) models trained without interruptions and the GNN-RL (3) models trained with interruptions. In general, especially for cases with $P_{interrupt} <$ 10%, on average, the GNN-RL schedulers perform better or at a similar level comparable to the SPT scheduler.

In summary, this experimental study with benchmark instances of different sizes shows that the GNN-RL methods remain robust for production scenarios with different job shop sizes even though the models are trained originally with the baseline benchmark of different size. More detailed tables of results on selected LA benchmarks of various sizes under a range of $P_{interrupt}$ values are reported in Supplemental Material.

## 5. SUMMARY AND DISCUSSION

The ability to assign jobs to machines under possible changes of operational conditions is important in practice. This study shows that GNN and RL can be a viable approach for solving the ISBJSSP, a complex and computational demanding problem subjected to unforeseen changes to the problem condition. In this work, we proposed a dynamic disjunctive graph formulation where nodes and edges are removed and reinstated during machine shutdowns and recoveries, respectively. Furthermore, we implemented a simulator to generate ISBJSSP instances for training and to validate the GNN-RL models for real-time scheduling of the ISBJSSP. The simulator accounts for interruption, swapping and blocking, which altogether adds complexity to the machine availability status and therefore affects the state and action space of the MDP formulation. For the simulations with no machine interruptions, the dispatching rule generated by the best trained GNN-RL scheduler achieves the best overall makespan exceeding that of the PDRs. (It should be noted that under perfect job shop conditions, mathematical optimization can produce more superior schedules [19].)

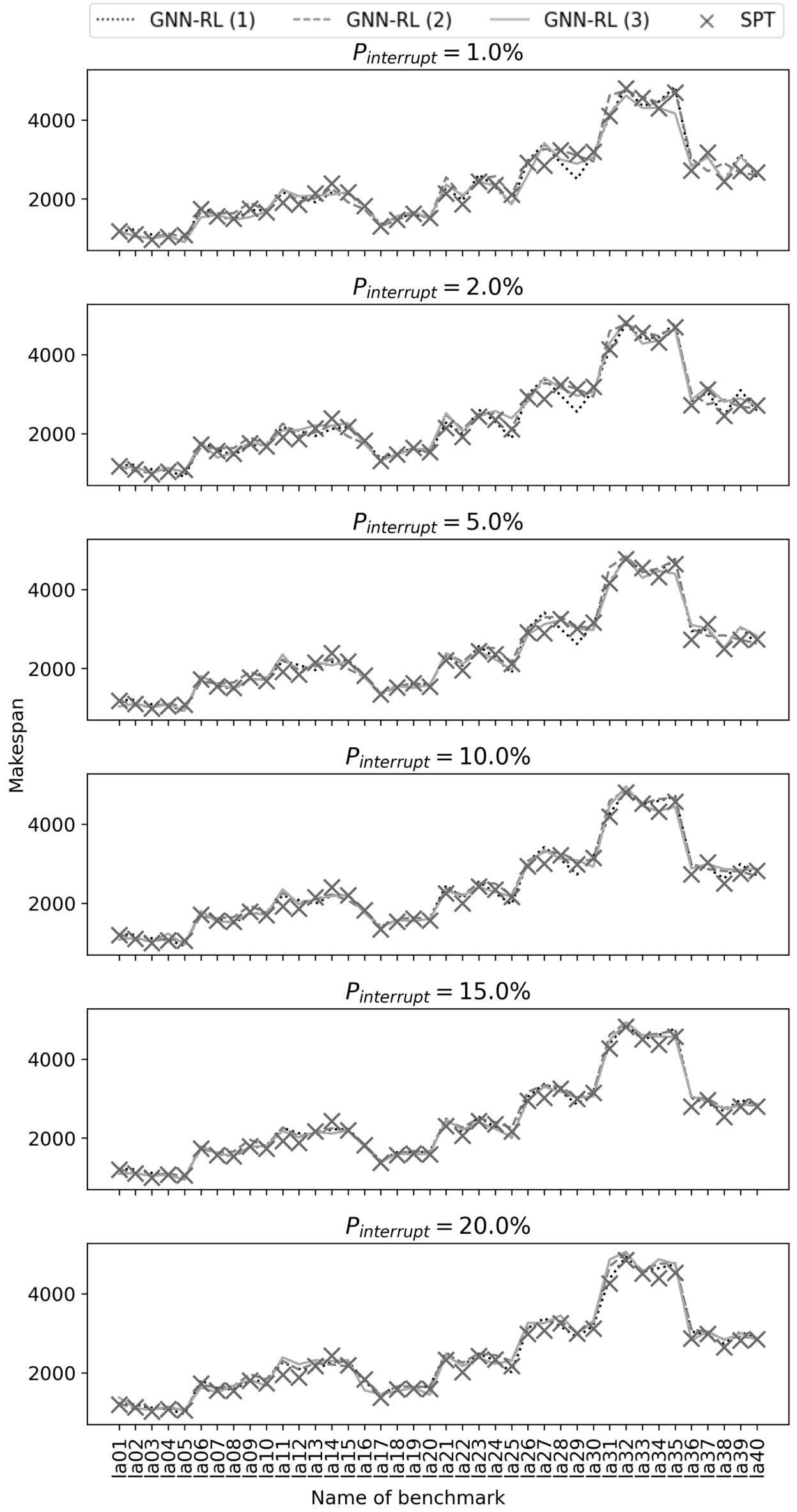

**Fig. 10 Mean of makespans for LA benchmarks.**

As the key objective of this study, we simulate scenarios where machines in the job shop can possibly be interrupted and shut down temporarily. The results show that the GNN-RL trained schedulers are robust under interruptions and outperform the PDRs approaches when the probability of machine interruption is low (less than 10% in the examples). In practice, it is very unlikely that the job shop would remain operational when the machines are deemed to have a high probability for being shut down. Furthermore, with an emphasis on robustness and practicality, our experimental study shows that the GNN-RL method is able to observe constantly changing disjunctive graph states and furthermore, to generalize to different job shop sizes subjected to a range of interruption probabilities. Given the speed of outputting actions and its decent performance, the GNN-RL method represents a viable approach applicable to real manufacturing problems that can be closely modeled as an ISBJSSP. While our research utilizes random simulations, domain-specific knowledge should be strategically incorporated in the real production environment to, for example, build specialized reward functions and fine-tune hyperparameters.

Future work could focus on developing methodologies for fine-tuning the parameters of the GNN and RL models in order to further improve the scheduling results. Empirical studies of other learning, adaptive algorithms could be explored. Due to the dynamic nature of the ISBJSSP problem, where the machine availability pattern changes and uncertainties are introduced, experimental studies are needed to further validate the simulator and the accompanying GNN-RL algorithm to demonstrate on machine interruptions in real world job shop environments. For instance, machine failures during less ideal timings, such as during an active operation, could significantly impact the overall job quality. Such type of interruption could therefore be investigated by future studies developing new simulators and scheduling strategies. Finally, further investigation on the proposed workflow may include other domain-specific constraints, such as limited buffer capacity, queue time constraints and multi-line scheduling that are commonly encountered in semiconductor manufacturing.

## ACKNOWLEDGEMENTS

The research was partially supported by Samsung Electronics Co. Ltd., Agreement Number SPO-168006, and the US National Institute of Standards and Technology (NIST), Grant Number 70NANB22H098, awarded to Stanford University. The research has also been partially supported by the Stanford Center at the Incheon Global Campus (SCIGC), which is sponsored in part by the Ministry of Trade, Industry, and Energy of the Republic of Korea and managed by the Incheon Free Economic Zone Authority. Certain commercial systems are identified in this article. Such identification does not imply recommendation or endorsement by Samsung, NIST or SCIGC; nor does it imply that the products identified are necessarily the best available for the purpose. Further, any opinions, findings, conclusions, or recommendations expressed in this material are those of the authors and do not necessarily reflect the views of Samsung, NIST, SCIGC or any other supporting U.S. Government or corporate organizations.